\documentclass[11pt]{article}

\usepackage[preprint]{acl}

\usepackage{times}
\usepackage{latexsym}

\usepackage[T1]{fontenc}

\usepackage[utf8]{inputenc}

\usepackage{microtype}

\usepackage{inconsolata}

\usepackage{graphicx}

\usepackage{hyperref}       %
\usepackage{url}            %
\usepackage{booktabs}       %
\usepackage{amsfonts}       %
\usepackage{nicefrac}       %
\usepackage{xspace}
\usepackage{lipsum}

\usepackage{enumitem}

\usepackage[table]{xcolor}
\usepackage[dvipsnames]{xcolor}

\definecolor{d3red}{HTML}{d62728}
\definecolor{d3green}{HTML}{2ca02c}

\newcommand{\red}[1]{\textcolor{d3red}{#1}}
\newcommand{\green}[1]{\textcolor{d3green}{#1}}

\newcommand{\codedist}{\textsc{CodeDistiller}\xspace}

\title{\raisebox{-0.3\height}{\includegraphics[height=1cm]{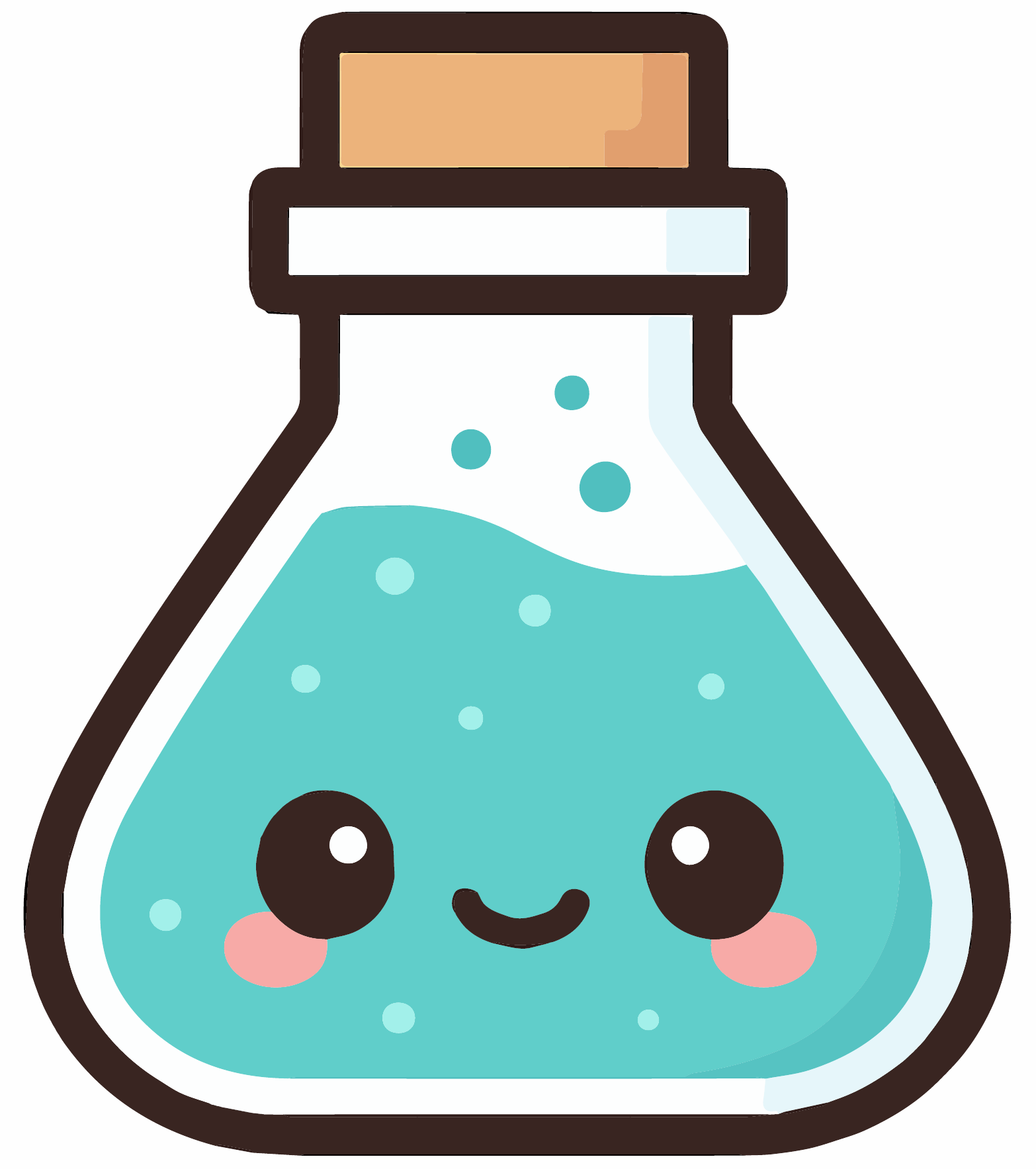}}
\textsc{CodeDistiller:} Automatically Generating Code Libraries for Scientific Coding Agents}

\author{
 \textbf{Peter A. Jansen\textsuperscript{1,2}},
 \textbf{Samiah Hassan\textsuperscript{1}},
 \textbf{Pragnya Narasimha\textsuperscript{1}}
\\
 \textsuperscript{1}University of Arizona,
 \textsuperscript{2}Allen Institute for Artificial Intelligence\\ 
 \texttt{pajansen@arizona.edu}
}

\begin{document}
\maketitle

\begin{abstract}
Automated Scientific Discovery (ASD) systems can help automatically generate and run code-based experiments, but their capabilities are limited by the code they can reliably generate from parametric knowledge alone. As a result, current systems either mutate a small number of manually-crafted experiment examples, or operate solely from parametric knowledge, limiting quality and reach. We introduce \codedist, a system that automatically distills large collections of scientific \textsc{Github} repositories into a vetted library of working domain-specific code examples, allowing ASD agents to expand their capabilities without manual effort.  Using a combination of automatic and domain-expert evaluation on 250 materials science repositories, we find the best model is capable of producing functional examples for 74\% of repositories, while our downstream evaluation shows an ASD agent augmented with a \codedist generated library produces more accurate, complete, and scientifically sound experiments than an agent with only general materials-science code examples. We also evaluate \textsc{LLM-as-a-judge} ratings against domain-expert ratings in an A/B testing paradigm, finding moderate agreement and suggesting that inexpensive proxy metrics may be feasible for evaluating scientific discovery systems at scale.\footnote{Video: \url{http://youtu.be/RQxSGFbGZSc}, Repository:\\ \url{www.github.com/cognitiveailab/codedistiller}}
\end{abstract}

\section{Introduction}

\begin{figure}[t!]
  \centering
  \includegraphics[scale=0.98]{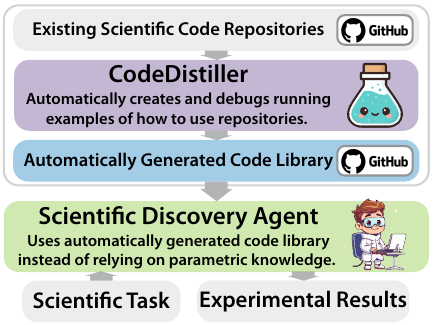}
  \caption{\footnotesize \codedist distills a large collection of \textsc{Github} repositories into a library of reusable scientific code, allowing \textsc{Code-RAG} style scientific discovery agents to perform tasks beyond their parametric knowledge.}

  \label{fig:small-overview}
\end{figure}

While automated scientific discovery has been explored for decades \cite{Simon1981ScientificDA}, particularly in the context of literature-based discovery \cite{Swanson1986FishOil,yang-etal-2024-large-language} and data-driven discovery \cite{LANGLEY1989283,bodhi2024datavoyager}, recent advances in code generation models \cite[e.g.][]{Chen2021EvaluatingLL,Jiang2024ASO} have spurred the development of experiment-driven discovery agents in domains such as computer science where experiments can be run primarily through code. Typically these agents are supplied with a research task, then must generate (and iteratively debug) code that runs a computational experiment before writing a report describing the results and conclusions, as shown in Figure~\ref{fig:small-overview}. These systems have recently been demonstrated to produce novel, if incremental, scientific discoveries \cite{Weng2025DeepScientistAF,Zhang2025NovelSeekWA,jansen-etal-2025-codescientist}.

A central challenge is that scientific experiments typically require highly specific methods, measurements, and protocols, whereas coding agents are limited to generating experiment code using whatever parametric knowledge they acquired during training. Existing automated scientific discovery agents therefore either mutate pre-existing experiment code \citep[e.g.][]{lu2024aiscientistfullyautomated} or require the experimenter to pre-generate a library of existing vetted code examples they can combine \citep[e.g.][]{jansen-etal-2025-codescientist}. In this work we address this limitation by building a system that can automatically build a library of functional domain-specific code examples at scale, allowing systems to automatically increase their domain-specific capabilities while reducing their reliance on manual code construction.

The contributions of this work are:
\begin{enumerate}[itemsep=0pt, topsep=2pt]
    \item \codedist, a system for automatically converting \textsc{Github} code repositories in specialized scientific domains into debugged, working examples suitable for incorporation in automated scientific discovery systems.
    \item An evaluation in the materials science domain, showing that the best and worst performing base models are capable of successfully distilling between 26\% to 74\% of 250 repositories, with different price/performance tradeoffs.
    \item A downstream evaluation showing that an automated discovery system augmented with a library of code automatically built with \codedist generates more accurate, complete, and scientifically sound output than a baseline model with only general materials science code examples.
    \item A detailed characterization of the agreement (or disparity) between \textsc{LLM-as-a-Judge} vs domain-expert judgments in the materials science domain, showing moderate agreement for evaluating downstream automated scientific discovery quality using A/B testing, but mixed agreement (depending on the model) when evaluating the code distillation task.
\end{enumerate}

\section{Related Work}

{\flushleft\textbf{Automated Scientific Discovery:}} Agents for automated scientific discovery can be divided into those focusing on literature-based discovery \citep[e.g.][]{Swanson1986FishOil,yang-etal-2024-large-language}, data-driven discovery \citep[e.g.][]{bodhi2024datavoyager,mitchener2025kosmosaiscientistautonomous}, and experiment-driven discovery—the latter being the focus of this work. Several agents make use of solely computational experiments for discovery. \textsc{AI Scientist} \cite{lu2024aiscientistfullyautomated} mutates the code of existing experiment repositories to create novel experiments. \textsc{AgentLab} \cite{Schmidgall2025AgentLU} generates experiment code after reviewing relevant literature. Some approaches rely on expert-generated tool interfaces to domain-specific tools, such as in chemistry \citep[\textsc{ChemCrow};][]{Bran2024Augmenting} or biology \citep[\textsc{Biomni};][]{Huang2025Biomni}. \codedist addresses the approach taken by \textsc{CodeScientist} \cite{jansen-etal-2025-codescientist}, where a \textsc{Code-RAG} agent retrieves multiple relevant code examples from a vetted code library that it can combine to build complex computational experiments, and reduce reliance on parametric knowledge. While the 6 discoveries made by \textsc{CodeScientist} relied on a code library built through a combination of manual generation and expert curation of LLM-generated examples, \codedist allows for augmenting \textsc{Code-RAG} libraries through automatically generated and vetted examples.

{\flushleft\textbf{Tasks:}} Several existing tasks are similar to the example distillation task. The \textsc{SUPER} benchmark \cite{bogin-etal-2024-super} evaluates agents' capacity to set up and replicate specific research results from a \textsc{Github} repository, with current best performance at 16\%. \textsc{RexBench}~\cite{edwards2025rex} and \textsc{TM-Bench}~\cite{wolflein-etal-2025-llm} demonstrate automatically repurposing 12 and 15 research repositories (respectively) from \textsc{Papers-With-Code} to new research tasks. \textsc{ENVBENCH} \cite{eliseeva2025envbench} addresses LLM configuration of development environments. \textsc{ResearchCodeBench} \cite{Hua2025ResearchCodeBenchBL} evaluates whether LLMs can implement 212 coding challenges drawn from 20 research articles using only parametric knowledge, finding the best model achieved less than 40\% success, supporting the notion that distilling examples from repositories may increase recall. \textsc{GISTIFY} \cite{lee2025gistifycodebaselevelunderstandingruntime} requires generating fully self-contained examples from 5 \textsc{Github} repositories whose output passes sets of predefined \textsc{pytests}.

In contrast, \codedist focuses on building a code library for \textsc{CodeRAG}-style automated discovery systems, which can require integrating code across multiple repositories. \codedist does not require repositories to contain unit tests, and is evaluated at greater scale than the above systems—both using automatic \textsc{LLM-as-a-judge} metrics \cite{zheng2023judging}, as well as human domain expert evaluation in our domain of interest (materials science).

\begin{figure*}[t!]
  \centering
  \includegraphics[scale=1.00]{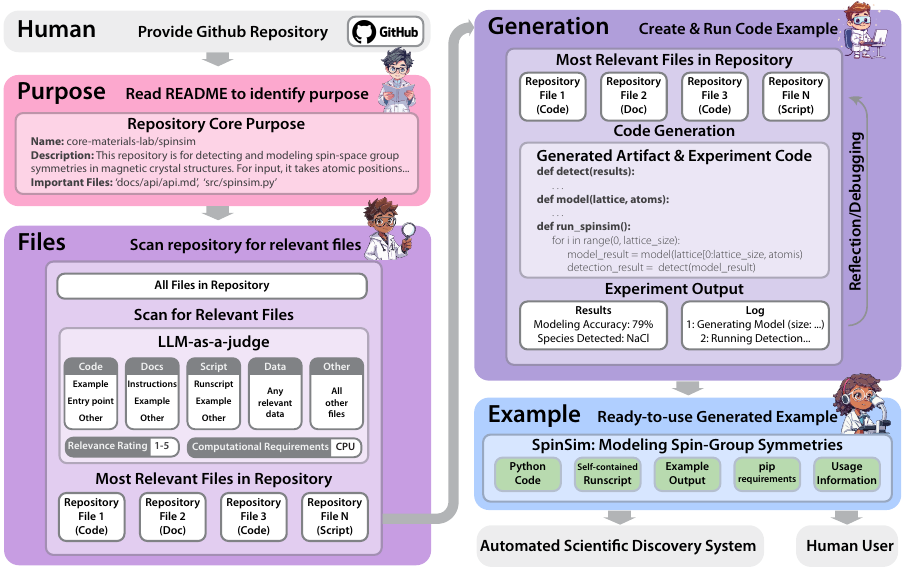}
  \vspace{-1mm}
  \caption{\footnotesize An overview of the core stages of the \codedist workflow, including identifying the core purpose of the repository, identifying files relevant for building an example, and the example generation and debugging process.}
  \label{fig:system_overview}
  \vspace{-1mm}
\end{figure*}

\begin{table*}[t]
\centering
\small
\setlength{\tabcolsep}{6pt}
\begin{tabular}{lccc}
\toprule
~ & \multicolumn{3}{c}{\textbf{Agent Base Model}} \\
\textbf{Metric} & \textbf{GPT-OSS-120B} & \textbf{GPT-5} & \textbf{Claude Sonnet 4-5} \\
\midrule
\rowcolor{gray!15}\multicolumn{4}{l}{\textbf{Overall Performance}}\\
\midrule
~~\textit{(Auto)} Successfully Completed (LLM-as-a-judge)   &   61.6\%    & 70.4\%        &   75.6\%    \\
~~\textit{(Manual)} Code Executed without Error$\dagger$             &   29.6\%     & 69.0\%    & 75.6\% \\
~~\textit{(Manual)} Demonstrates Repository Functionality$\dagger$   &   29.6\%     & 69.0\%    & 75.6\% \\
~~\textit{(Manual)} Correct Functionality$\dagger$                   &   25.9\%     & 60.5\%    & 74.1\%        \\

\midrule
\rowcolor{gray!15}\multicolumn{4}{l}{\textbf{Runtimes}}\\
\midrule
~~Avg. Runtime (successful cases)         &   13.8 mins    &   20.3 mins   &   19.0 mins   \\   
~~Avg. Runtime (unsuccessful cases)       &   16.3 mins    &   47.0 mins   &   21.9 mins   \\
~~Avg. Debug Iterations (successful cases)       &   2.4         & 2.2 &    1.9    \\
\midrule
\rowcolor{gray!15}\multicolumn{4}{l}{\textbf{Costs}}\\
\midrule
~~Avg. Cost (successful cases)       &   \$0.09    &   \$0.70    &   \$1.71    \\
~~Avg. Cost (unsuccessful cases)     &   \$0.13    &   \$1.68    &   \$2.26    \\
\bottomrule
\end{tabular}
\caption{Summary statistics of \textsc{CodeDistiller} performance as a function of using different base models. $\dagger$ Manual analysis was completed by a domain expert on a subsample of 50 repositories that were marked as \textit{successfully completed} by the agent, and normalized to have a common denominator with automatic results.}
\label{tab:performance-by-model}
\end{table*}

\section{System Overview}

The \codedist agent workflow is shown in Figure~\ref{fig:system_overview}.  First, large-scale static information gathering steps search and identify code, documentation, and other files relevant to understand the repository and identify highly relevant files. Then, those highly-relevant files are used in a dynamic workflow to generate and iteratively debug a working example of the core repository functionality.

\subsection{Relevant File Classification}

Repositories can contain hundreds of nested files, but only a small subset of those files -- such as top-level APIs, documentation, or existing examples (if available) -- tend to be relevant for building example code that demonstrates a repository's core functionality.  Similarly, as a pragmatic consideration, the base language model used for generating and debugging the code example has a limited context window, and by providing only the most task-relevant files in the context, we increase the likelihood of success.

To this end, each file in a repository is individually provided to a prompt, which classifies the file contents into several coarse-grained types (shown in Figure~\ref{fig:system_overview}), such as \textit{code, documentation, scripts, data, or other}. Three of the classes (code, documentation, and scripts) are further subdivided into finer-grained categories (e.g., \textit{existing examples}, \textit{instructions}, \textit{entry points}) to help determine whether a file is likely to have utility. In addition to classification, this scanning procedure also assigns an overall relevance score for each file (on a scale of 1–5) and generates complementary metadata, such as whether the file mentions special computational requirements (e.g., \textsc{GPUs}) or contains critical task-relevant information (e.g., configuration instructions) needed by the code-generation tool.

\subsection{Code Generation}

The code generation step provides the highest-ranked files and the repository's core purpose to a code generation system, whose task is to build, execute, and iteratively debug a working example of the repository's core functionality. First, a prompt (including the most relevant files) is provided to a base LLM, which generates four components: (1) executable \textsc{Python} code, (2) \textit{Python} library requirements, (3) a \textsc{bash} runscript with \textsc{Conda} environment setup, and (4) associated metadata, including a description; inclusion/exclusion criteria (i.e., what purposes the example is suited for or inappropriate for); computational resource requirements such as CPU cores, GPUs, RAM, and disk; and whether the repository's code is stand-alone or requires user interaction. This functionality is provided by a modified version of the \textsc{CodeScientist} automated scientific discovery agent, adapted for the example-distillation task, as \textsc{CodeScientist} includes detailed instrumentation for iteratively debugging LLM-generated code.

Once an initial example is generated, the code is automatically run in an \textsc{Ubuntu} cloud container, and its output instrumented and captured. The output includes debugging-oriented instrumentation (such as a timestamped log file recording each major operation, an overall \textsc{JSON} results file, and raw \textsc{stdout/stderr} logs). The code is also encouraged to generate figures and other human-readable demonstrations useful for verifying functionality. When execution is complete, or exceeds preset runtime thresholds, the output is provided to an \textsc{LLM-as-a-judge} prompt that determines whether the code functioned correctly or contains issues. If issues are found, the code is reflected -- using the current version and execution logs as input -- and the modified code is executed. This process repeats until the LLM judges the example to have run successfully, at which point the example generation process ends and the example is provided to the user (or ASD system). To control for costs and runtime, if debugging continues for too many iterations (typically 8), the process is terminated and the example is marked as unsuccessful.

\section{Evaluation}
\label{sec:evaluation}

\subsection{Materials Science Repositories}

We evaluate the performance of \codedist in a materials science use case, where \textsc{Github} repositories in the materials science domain are provided, and \codedist must distill examples of using those repositories suitable for downstream human or automated scientific discovery agent use. 
 
{\flushleft\textbf{Materials Science Libraries:}} A materials science subject matter expert assembled a list of 30 popular materials science libraries in \textsc{Python}, including \textsc{PyMatgen} (a materials analysis library), \textsc{ASE} (tools for atomic simulation), \textsc{LAMMPS} (a molecular dynamics simulator), and \textsc{PyCalphad} (a computational thermodynamics modeler).

{\flushleft\textbf{Github Repositories:}} Using the \textsc{Github API}, we identified all active \textsc{Github} repositories whose source files mentioned at least one of the list of materials science libraries, and whose repository was licensed under a permissive open source license.  This resulted in 3,802 unique repositories.  To minimize cost, for this evaluation we randomly subsampled this to a list of 250 repositories. 

\subsection{Evaluation Metrics}

We evaluate the performance of \codedist on three main sets of metrics: automatic assessments of task performance, manual assessments completed by a materials science domain expert, and summary statistics of costs and runtimes. 

{\flushleft\textbf{Successfully Completed (Auto):}} The LLM-as-a-judge \cite{zheng2023judging} that makes a binary assessment of whether \codedist has executed correctly and faithfully to the task purpose, and marks the example as complete.

{\flushleft\textbf{Code Executed without Error (Domain Expert):}} As above, but a domain expert manually inspects the output of the repository to verify that the code executed, and that log files are present. 

{\flushleft\textbf{Demonstrates Repository Functionality (Domain Expert):}} A  domain expert examines the final generated code, and makes a binary judgment as to whether the code does demonstrate the core functionality of the input \textsc{Github} repository.  For example, if the input repository is for a molecular dynamics simulator, the generated example must perform a molecular dynamics simulation, rather than simply generating plots, configuration files, or other non-core functionality. 

{\flushleft\textbf{Correct Functionality (Domain Expert):}} A domain expert examines the output of the code, and makes a binary assessment as to whether the output appears correct and faithful to the expected output.

\subsection{Experiments}

{\flushleft\textbf{Models:}} We evaluated \codedist using three popular LLM base models that represent different price/performance points: \textsc{GPT-OSS-120B}, a popular open-weight model, \textsc{GPT-5}, a popular reasoning model that includes web search tool calls, and \textsc{Claude Sonnet 4.5}, a popular reasoning model for code generation. The \codedist pipeline can use different models for the file classification step (which can make use of inexpensive, fast models) and the code generation and debugging steps (which requires more capable models).  As such, the file classification step used the less-expensive version of each model family (e.g. \textsc{GPT-5-mini}, \textsc{Claude Haiku 4.5}), with the exception of \textsc{GPT-OSS-120B}, which was both fast and inexpensive enough to use for all parts of the pipeline.

{\flushleft\textbf{Domain Expert Evaluation:}} A subset of 50 repositories were selected for manual evaluation by the domain expert, based on the criterion that they each were marked as successful by each of the 3 base model's automated \textit{successfully completed} metric.  The domain expert has a graduate degree in materials science. 

\subsection{Results}

The results are shown in Table~\ref{tab:performance-by-model}. Overall, automatic measures of success increase with model cost and complexity, ranging from 62\% of repositories being marked as successfully having an example generated by \textsc{GPT-OSS-120B}, up to 76\% by \textsc{Claude Sonnet 4.5}. This increase in performance comes with a steep increase in cost -- with the average \textsc{Claude}-generated example costing 19 times more than the \textsc{GPT-OSS} version.

{\flushleft\textbf{Overcounting:}} Manual analysis shows a significant mismatch between automatic \textsc{LLM-as-a-judge} and manually-measured performance, with each model overestimating its performance.  This is most stark for \textsc{GPT-OSS-120B}, which reports that it has successfully made examples for 62\% of repositories, while the domain expert evaluation suggests the actual proportion with correct functionality is 26\%.  This undercounting is less for \textsc{GPT-5}, whose manually-validated performance decreases from 70\% to 61\%, and \textsc{Claude 4.5}, that decreases modestly from 76\% to 74\%. 

{\flushleft\textbf{Resource Cost:}} Overall, the models share similar runtimes, and similar numbers of debug iterations.  In this materials science domain, the example generation process tends to take between 13 and 21 minutes for successful examples, while taking between 1.9 and 2.4 debug iterations, on average.  Unsuccessful attempts are a significant resource cost in terms of both LLM \textsc{API} cost and time.  This is particularly true in that the model tends to continue to iterate the code until it hits a hard limit on the number of debug iterations, which necessarily costs more than successful examples, which tend to complete after an average of only two to three debug iterations.  Here we mitigate this cost by limiting to 8 debug iterations, such as not to incur large cost in the face of diminishing returns as the number of debug iterations increases.

\begin{figure}[t!]
  \centering
  \includegraphics[scale=0.925]{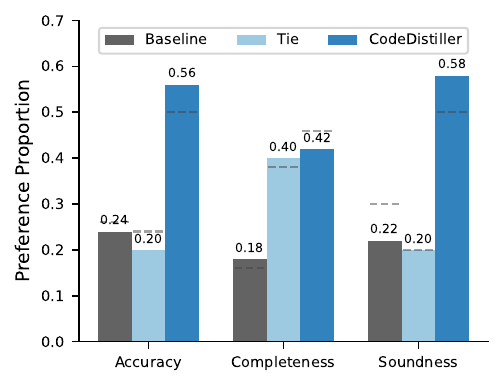}
  \vspace{-8mm}
  \caption{\footnotesize Results of A/B testing, showing the proportion of times the domain expert \textit{(main results)} or LLM-as-a-judge model \textit{(dashed lines)} preferred the experimental output from the baseline model (with generic materials science code examples) versus the model augmented with a \codedist-generated library.   Values represent the average of 50 experimental tasks implemented using \textsc{CodeScientist}.}
  \vspace{-3mm}
  \label{fig:ab_test}  
\end{figure}

\section{Downstream Evaluation}
Here we evaluate whether incorporating automatically generated code examples from \codedist improves the performance of a downstream automated scientific discovery system.

\begin{table}[t]
\centering
\footnotesize
\setlength{\tabcolsep}{6pt}
\begin{tabular}{p{0.95\linewidth}}
\toprule

\multicolumn{1}{c}{\textbf{Better Data}} \\
\midrule
\textbf{Solution 1 [Baseline]} uses \red{synthetic fabricated data (20 hand-picked molecules replicated) instead of actual Tox21, producing predictions with no scientific validity or real-world grounding}.\\
\textbf{Solution 2 [CodeDistiller]} uses the \green{actual Tox21 dataset with 6,258 real compounds across 12 toxicity assays, producing scientifically grounded predictions from properly trained models.} \\
\midrule
\multicolumn{1}{c}{\textbf{Better Modeling}} \\
\midrule
\textbf{Solution 1 [Baseline]} uses a \red{generic Lennard-Jones potential not parameterized for Ge, Sb, or Te, resulting in extreme, unphysical volume collapses (80-93\%) that do not represent accurate materials behavior.} \\
\textbf{Solution 2 [CodeDistiller]} \green{uses CHGNet, a materials-specific ML potential trained on DFT data, producing physically reasonable volume changes (-16\% to +75\%) that reflect realistic structural relaxation behavior.} \\
\midrule
\multicolumn{1}{c}{\textbf{More Canonical Solutions}} \\
\midrule
\textbf{Solution 1 [Baseline]} provides numerical values with proper units for all parameters, but \red{implements calculations from scratch with a manual database, leading to some discrepancies in atomic size difference values (e.g., AlTiVNb \u03b4 = 3.60\% vs 5.428\% in Solution 2).} \\
\textbf{Solution 2 [CodeDistiller]} \green{uses established computational materials science libraries (pymatgen and Parameter-Calculator-for-CCA) that are peer-reviewed and widely validated, likely providing more reliable atomic property data and calculation implementations.}\\
\bottomrule
\end{tabular}
\caption{Example A/B test preference ratings, highlighting more accurate science from \codedist.}
\label{tab:ab-test-examples}
\end{table}

\subsection{Task Framing}
To understand whether incorporating \codedist-generated examples into a \textsc{Code-RAG} agent would improve performance, we constructed a set of materials science tasks centered specifically around the primary function of a subset of the materials science repositories generated in Section~\ref{sec:evaluation}.  This includes tasks on magnetic symmetry, cheminformatics, condensed matter, and other related tasks.  An automated scientific discovery agent ran with and without the code examples, and the output was compared.

{\flushleft\textbf{Tasks:}} A diverse set of 60 discovery problems targeted at the primary purpose of the code examples were generated using \textsc{Claude Sonnet 4.5}. The list of repositories was filtered by the domain expert to a set of 12 with minimal external data requirements, as materials science tasks can often require closed-source data that is difficult to acquire, and 5 problems were generated for each repository.

{\flushleft\textbf{Discovery Agent:}} The tasks were provided to \textsc{CodeScientist} in two configurations.  In the \textit{baseline} configuration, \textsc{CodeScientist} was given access to general materials-science domain code examples, including popular simulation packages for materials analysis, atomic simulation, molecular dynamics, and computational thermodynamics, as well as associated openly accessible data.  In the \textit{experimental} configuration, the set of code examples was expanded to include the relevant \codedist generated example appropriate for a given task. We used \textsc{Claude Sonnet 4.5} as a base model.  Each \textsc{CodeScientist} run was given a maximum of 15 debug iterations, 6 hours of total runtime (capped at 60 minutes per debug iteration), and up to \$5 USD of \textsc{LLM}-associated costs.  Because automated discovery is still challenging and low-recall, we submitted each of the baseline and experimental tasks twice, and for a given discovery problem, used the first run that successfully completed (if any) in the subsequent evaluation.

{\flushleft\textbf{Expert and Automated Evaluation:}} To measure whether \codedist examples provide benefit over the baseline system using only generic materials-science domain examples combined with parametric knowledge, we evaluated output using an A/B testing paradigm \cite{kohavi2020trustworthy,fisher1935design}. Specifically, for a given discovery problem, we provided the generated \textit{code}, \textit{raw experimental results}, and \textit{experiment report} from both models, and asked the \textit{domain expert} (manual condition) and an \textsc{LLM-as-a-judge} (automatic condition) to rate which performed better on three dimensions: \textit{accuracy}, \textit{completeness}, and \textit{soundness}. The output of each model was rated blind, and for LLM-as-a-judge each discovery problem was rated twice, counterbalancing presentation order \cite{wang-etal-2024-large-language-models-fair,zheng2023judging}. The A/B test allowed three ratings: a preference for A, B, or a tie, and was framed in a chain-of-thought paradigm requiring independent textual evaluations of each system before the final A/B rating. To control for problem difficulty, we only examined problems where both \textit{baseline} and \textit{experimental} models produced a solution (50 problems).

\begin{table}[t]
\centering
\small
\setlength{\tabcolsep}{6pt}
\begin{tabular}{lc}
\toprule
\textbf{Rating Dimension} & \textbf{Cohen's $\kappa$} \\
\midrule
Accuracy & 0.77 \\
Soundness & 0.70 \\
Completeness & 0.62 \\
\bottomrule
\end{tabular}
\caption{Agreement between manual (domain-expert) and LLM-as-a-judge A/B preference ratings across each rating dimension for the downstream discovery task.}
\label{tab:llm-judge-agreement}
\end{table}

\subsection{Downstream Results}
The results of the A/B test are shown in Figure~\ref{fig:ab_test}. Overall across metrics, in both domain-expert and LLM-as-a-judge conditions, the results show a strong preference for the system that includes \codedist generated examples.  While the baseline system is preferred in between 18\% to 24\% of cases across \textit{accuracy}, \textit{completeness}, and \textit{soundness}, the \textsc{CodeScientist} system augmented with \codedist is generally preferred in more than  half of cases, while approximately a quarter of problems resulted in ties.  Example reasoning traces for these preferences are shown in Table~\ref{tab:ab-test-examples}, highlighting the utility of automatically including vetted community-generated \textsc{Github} repositories for scientific computation over the parametric knowledge stored within the base model. 

{\flushleft\textbf{Expert and LLM-as-a-judge Agreement:}} Table~\ref{tab:llm-judge-agreement} reports interannotator agreement between manual (domain-expert) and LLM-as-a-judge ratings on the A/B task, measured using Cohen's Kappa \cite{cohen1960coefficient}, broken down by each of the three dimensions (\textit{accuracy}, \textit{completeness}, and \textit{soundness}). 
Overall results range between a Kappa of 0.62 and 0.77, which is interpreted as overall \textit{moderate} agreement \cite{mchugh2012interrater} -- though we note that the \textit{accuracy} dimension approaches \textit{strong} agreement, while the \textit{completeness} dimension borders being classified as \textit{weak} agreement.  This suggests that for \textit{accuracy} and \textit{soundness} evaluation dimensions, a strong LLM-as-a-judge model may serve as an approximate inexpensive proxy metric for expert ratings in this paradigm, as expert ratings were expensive to collect, requiring several weeks to examine reports.  It also suggests that a domain expert is still particularly important when evaluating the \textit{completeness} of experimental reports.

\section{Conclusion}
We present \codedist, a system for automatically converting scientific \textsc{Github} repositories into ready-to-use examples of core domain-specific functionality for code-based scientific discovery agents.  Through automated and manual evaluation in the materials science domain, we empirically demonstrate that the best base model can achieve 74\% performance on this distillation task, while a system using these automatically constructed code examples outperforms a baseline system in the accuracy, completeness, and scientific soundness of the output.  Our system is available as open source software.

\section*{Limitations}

{\flushleft\textbf{Domain expert:}} It is difficult and time consuming to measure whether the generated example is genuinely correct -- at the extreme, this would require the domain expert to generate a large set of tests, use the example code on several known problems, and verify the results match the literature.  Here, the domain expert examines the code, results, and any generated data/figures, and makes a good-faith effort to judge whether the code appears to be executing correctly and faithfully to the main function of the repository.  As such, while we use a domain-expert evaluation, it should be considered a time-restricted proxy evaluation rather than an exhaustive evaluation.

{\flushleft\textbf{Automatically identifying materials science repositories:}} In this work we automatically identify materials science repositories on \textsc{Github} by searching for repositories that contain at least one file with at least one import from a list of common materials science libraries.  We estimate, based on our manual analysis, that approximately half of the repositories collected in this way are directly related to materials science, while the remainder are unrelated, but happened to import (for example) a data analysis library used in materials science.  When conducting the manual analyses of \codedist performance, we preferentially selected repositories that were directly related to materials science for the domain expert to evaluate.

{\flushleft\textbf{Multi-domain evaluation:}} This work examines a materials science use case, which is a high-impact domain that makes use of significant computational experimentation.  Some of this computational tooling is open source and potentially available for use by the system, while other tooling is closed and, in the experiment framing described here, unavailable.  Similarly, tooling in materials science frequently relies on associated materials data, with similar challenges around open versus closed data availability.  If transferred to another domain, domain-specific challenges (including, but not limite to issues of software and data availability) may affect overall code distillation performance, as well as performance on downstream discovery tasks.

{\flushleft\textbf{Purpose-built versus general-purpose agents:}} A contemporary question in agent-based automated scientific discovery is whether purpose-built agents (such as \codedist) or general-purpose agents (such as Claude Code, Codex, and related systems) are better for specific tasks. There are many challenges in resolving this question -- systems have different input, budgets, models, architectures, ablations, and highly variable output -- making evaluating this in a controlled manner difficult, particularly in the present case where substantial manual evaluation by a domain expert is required.  This work is agnostic to this open question of architecture, and (depending on the system or context it is used in), \codedist could be used both as a purpose-built agent for pre-generating code libraries as a preprocessing step, but also as a tool or subagent that more general agents could call during their execution.

\section*{Acknowledgments}

This research was developed with funding from the Defense Advanced Research Projects Agency’s (DARPA) SciFy program (Agreement No. HR00112520300) to PJ at the University of Arizona. The views expressed are those of the author and do not reflect the official policy or position of the Department of Defense or the U.S. Government. This work was supported in part a Modal for Academics compute grant. PJ has an outside interest in the Allen Institute for Artificial Intelligence. This interest has been disclosed to the University of Arizona and reviewed in accordance with its conflict of interest policies. We thank the members of the DARPA Scientific Feasibility (SciFy) program and Peter Clark for thoughtful discussions.

\bibliography{anthology,custom}

\end{document}